\PassOptionsToPackage{hyphens}{url}
\documentclass[10pt, conference, compsocconf]{IEEEtran}

\usepackage{latexsym, amsmath, amsfonts, amsthm, bm, bbm, nicefrac, stmaryrd, thmtools, enumitem, centernot}
\usepackage{url, graphicx, xcolor, rotating}
\definecolor{svlinks}{rgb}{.0,0.3,0.6}
\usepackage[hang]{footmisc}
\usepackage[bookmarks,colorlinks=true,pdfhighlight=/O,linkcolor=svlinks,urlcolor=svlinks,citecolor=svlinks,
           pdftitle={Delete My Account: Impact of Data Deletion on Machine Learning Classifiers},
           pdfauthor={Tobias Dam, Maximilian Henzl, Lukas Daniel Klausner},
           pdfsubject={},
           pdfkeywords={}]{hyperref}
\usepackage[T2A,T1]{fontenc}
\usepackage[utf8]{inputenc}
\usepackage[russian,greek,UKenglish]{babel}

\usepackage{epsfig, listings}
\usepackage{multirow}
\usepackage{float}
\usepackage[section]{placeins}
\usepackage[all]{nowidow}
\usepackage{fnpct}
\usepackage{soul}

\usepackage{tikz, wrapfig, caption}
\usetikzlibrary{matrix}
\usepackage[normalem]{ulem}
\usepackage{booktabs}
\usepackage{array}
\usepackage[subfolder]{gnuplottex}
\usepackage{xfrac}
\usepackage{float}

\usepackage{placeins}
\usepackage[all]{nowidow}
\usepackage{fnpct}
\usepackage[capitalise,noabbrev]{cleveref}

\setlist{nosep}

\usepackage{etoolbox}
\pretocmd{\eqref}{Eq.~}{}{}

\begin{document}

\title{Delete My Account: Impact of Data Deletion on~Machine Learning Classifiers}

\author{
\IEEEauthorblockN{Tobias Dam}
\IEEEauthorblockA{
Institute of IT Security Research\\
St.\ P\"olten UAS\\
St.\ P\"olten, Austria\\
tobias.dam@fhstp.ac.at}
\and
\IEEEauthorblockN{Maximilian Henzl}
\IEEEauthorblockA{
Institute of IT Security Research\\
St.\ P\"olten UAS\\
St.\ P\"olten, Austria\\
is201849@fhstp.ac.at}
\and
\IEEEauthorblockN{Lukas Daniel Klausner}
\IEEEauthorblockA{
Institute of IT Security Research\\
St.\ P\"olten UAS\\
St.\ P\"olten, Austria\\
mail@l17r.eu}
}







\maketitle

\begin{abstract}
Users are more aware than ever of the importance of their own data, thanks to reports about security breaches and leaks of private, often sensitive data in recent years. Additionally, the GDPR has been in effect in the European Union for over three years and many people have encountered its effects in one way or another. Consequently, more and more users are actively protecting their personal data. One way to do this is to make of the right to erasure guaranteed in the GDPR, which has potential implications for a number of different fields, such as big data and machine learning.

Our paper presents an in-depth analysis about the impact of the use of the right to erasure on the performance of machine learning models on classification tasks. We conduct various experiments utilising different datasets as well as different machine learning algorithms to analyse a variety of deletion behaviour scenarios. Due to the lack of credible data on actual user behaviour, we make reasonable assumptions for various deletion modes and biases and provide insight into the effects of different plausible scenarios for right to erasure usage on data quality of machine learning. Our results show that the impact depends strongly on the amount of data deleted, the particular characteristics of the dataset and the bias chosen for deletion and assumptions on user behaviour.

\end{abstract}


\section{Introduction}

The General Data Protection Regulation (GDPR)~\cite{gdpr}, in effect since 25 May 2018, was a significant development for improving privacy and protecting the data of EU citizens. Among various rights granted to citizens regarding their data (such as the \emph{right of access by the data subject}), the GDPR also grants the \emph{right to erasure}. The right to erasure, also called the \emph{right to be forgotten}, empowers citizens to request the deletion of their personal data, with only a limited set of explicitly defined exceptions.

Personal data as defined in the GDPR (as ``information relating to a natural person'') includes identifiers, such as names, but also characteristics that may indirectly identify a person, such as data on their economical or social identity. In recent years, the amount of personal information that is collected, stored and processed has been steadily increasing. More and more companies apply concepts like big data~\cite{enisa-bigdata, parliament-bigdata} as well as machine learning in order to automatically process large amounts of collected data in order to provide personalised services, sell targeted advertising or conduct research.

As privacy awareness is continuously increasing over the past few years and the \emph{right to erasure} becomes more well-known, people are more likely to make use of their rights and request the deletion of their data. By way of example, the announcement of a modification to the privacy policy of WhatsApp in January 2021~\cite{nicas_2021} caused millions of users to switch to other messenger services. It stands to reason that at least part of these users also deleted their accounts and made use of their right to erasure.

Such increases of the usage rates of the right to erasure and the resulting loss of data in datasets used for machine learning tasks might impact the quality of those results. So far, little research has been conducted on the impact of the use of the right to erasure on machine learning. The primary aim of this paper is to investigate the influence of different amounts and distributions of data removed from datasets on the performance of various classification algorithms. We define a number of scenarios with different likelihoods to make use of the right to erasure based on demographic and socioeconomic features. We performed several experiments utilising these scenarios and different percentages of deleted data records and provide an in-depth analysis about the impact of the right to erasure on machine learning results.

In particular, the main contributions of this paper are:
\begin{itemize}
    \item We present an in-depth analysis of the impact of the right to erasure applied according to different deletion likelihood scenarios on machine learning classification performance.
    \item We compare our results across several datasets and ML algorithms.
\end{itemize}

The remainder of this paper is structured as follows: We discuss related work in \autoref{sec:relatedwork} and give an overview of the ML algorithms as well as the basis for the deletion scenarios used in our study in \autoref{sec:methodology}. We present the experimental setup along with our deletion methods, deletion modes and utilised datasets in \autoref{sec:design}. In \autoref{sec:evaluation}, we evaluate our results and present our analysis. Possible future work is detailed in \autoref{future work} and \autoref{sec:conclusion} concludes the paper.
\section{Related Work}
\label{sec:relatedwork}

While our work is associated with the growing research on data deletion in machine learning, relatively few studies have been conducted so far that focus on analysing the performance impact of the right to erasure. Existing studies (also often referencing the GDPR) mostly explore efficient deletion methods for ML models without the need to re-train from scratch, such as~\cite{bourtoule2020unlearning, ginart2019datadeletion, izzo2021approxdatadeletion}.

A study comparable to ours in regards to analysing the performance impact is Malle et al.~\cite{malle2016righttobeforgotten}. Their methodology is similar to our approach, but we analyse more datasets and use multiple different deletion methods.
\section{Methodology}
\label{sec:methodology}

In the following sections we describe the approaches investigated and compared in our study. The approaches include the ML techniques described below.

\subsection{Machine Learning Algorithms}
For the present study, we investigated a selection of popular supervised machine learning methods, namely Support Vector Machines~(SVM), $k$-Nearest Neighbour~($k$-NN), Random Forests~(RF) and Extreme Gradient Boosting~(XGBoost).\footnote{Since we are using precisely the same algorithms as in our previous work~\cite{slijepcevic2021kanonymity} 
, we reuse the algorithms' descriptions from that paper in this subsection.}

\textbf{SVMs} are popular supervised machine learning methods used for classification and regression. SVMs are effective and robust for high-dimensional input data (in many cases even if the number of features is greater than the number of samples). They are also versatile and flexible, as many kernel functions (e.\,g.\ linear, polynomial, radial basis function) can be specified as decision functions, thus allowing high adaptability to the input data. In contrast to more complex kernels, linear kernels are characterised by significantly shorter runtimes and little overfitting while still yielding comparatively good results. SVMs are sensitive to hyperparameters, e.\,g.\ the cost parameter $C$ in the case of the linear kernel.

\textbf{$k$-NN} is a simple and intuitive instance-based algorithm that does not require any actual model training process. To determine the class for a given test sample, a majority decision is made based on the class membership of a given number of nearest neighbours from the training data. To determine which neighbours are closest, a similarity metric (e\,.g.\ Euclidean distance) is used. For a sufficiently large dataset, great results can be obtained, but with unbalanced data the algorithm encounters difficulties. The method is very sensitive to the number of neighbours used to determine the class of a tested instance.

\textbf{RFs}~\cite{breiman2001random} are robust supervised machine learning methods based on an ensemble of simple decision trees. Individual decision trees are relatively inflexible and not robust because even small changes in the data can cause the generated decision trees to look very different. Building RFs involves first generating simple decision trees from different subsamples of the data and then combining the results (e.\,g.\ by averaging their probabilistic predictions) into a relatively robust joint model. RFs are sensitive to the number of decision trees employed, i.\,e.\ a sufficient number of decision trees is necessary to obtain robust predictions.

\textbf{XGBoost}~\cite{chen2016xgboost} is a highly optimised and efficient variant of Gradient Boosting~\cite{friedman2001greedy}. Similar to RFs, Gradient Boosting combines a set of decision trees to provide robust predictions. The difference lies in the generation of decision trees: RFs generate independent decision trees on random subsets of data and then combine their results, whereas Gradient Boosting learns the trees iteratively and also learns from existing trees.

\subsection{Right to Erasure and Privacy Awareness}
\label{sub:Right to Erasure and Privacy Awareness}
Unfortunately, despite the fact that the GDPR has been in force for over three years now (since 25 May 2018), there still seems to be no existing research literature on how the likelihood of using the right to erasure correlates with different demographic and socioeconomic features. Several hypotheses seem to make sense \emph{prima facie}; to give just a few examples:
\begin{itemize}
    \item Women and members of marginalised groups (such as People of Colour, members of the queer community, disabled or neurodivergent people, \dots) might be more likely to make use of the right to erasure as a form of digital self-defence against datafication, harassment, etc.
    \item People with more means, access and power (i.\,e.\ people from higher income percentiles or of higher socioeconomic status) might be more likely to make use of the right to erasure, both because they might be more aware of their rights as well as because they might feel they have more to hide.
    \item More competent users of technology might be more likely to use the right to erasure, as they are more likely to know about the possible negative effects of datafication.
\end{itemize}
In lieu of studies to base our experiments on, we can only make plausible assumptions and test different conceivable distributions of deletion likelihood.

\section{Study Design}
\label{sec:design}



\begin{table*}
    \centering
    \begin{tabular}{|m{6cm}|c|m{5cm}|m{4cm}|}
        \hline
        \multicolumn{1}{|c|}{\textbf{dataset}} & \multicolumn{1}{|c|}{\textbf{entries}} & \multicolumn{1}{|c|}{\textbf{QIDs}} & \multicolumn{1}{|c|}{\textbf{target variable}} \\
        \hline
        \textsc{adult}: \emph{Adult Dataset} (or \emph{Census Income Dataset}) & 30,162 & \texttt{sex}, \texttt{age}, \texttt{race}, \texttt{marital-status}, \texttt{education}, \texttt{native-country}, \texttt{workclass}, \texttt{occupation} & \texttt{salary-class}: \texttt{<=50K} or \texttt{>50K} \\
        \hline
        \textsc{cahousing}: \emph{California Housing Prices Dataset} & 20,640 & \texttt{housing\_median\_age}, \texttt{median\_house\_value}, \texttt{median\_income}, \texttt{longitude}, \texttt{latitude} & \texttt{ocean\_proximity}: \texttt{<1H OCEAN}, \texttt{INLAND} or \texttt{NEAR OCEAN} \\
        \hline
        \textsc{cmc}: \emph{Contraceptive Method Choice Dataset} & 1,473 & \texttt{wife\_age}, \texttt{wife\_edu}, \texttt{num\_children} & \texttt{contraceptive\_method}: \texttt{no\_use}, \texttt{long-term} or \texttt{short-term} \\
        \hline
        \textsc{mgm}: \emph{Mammographic Mass Dataset} & 830 & \texttt{bi\_rads\_assessment}, \texttt{age}, \texttt{shape}, \texttt{margin}, \texttt{density} & \texttt{severity}: \texttt{benign} or \texttt{malignant} \\
        \hline
    \end{tabular}
    \vspace{0.5em}
    \caption{Overview of the main characteristics of the datasets used in our experiments.}
    \vspace{-1.5em}
    \label{table:datasets}
\end{table*}

In order to perform our comparative analysis, we apply an approach similar to the one in our previous study~\cite{slijepcevic2021kanonymity}.
We implement different methods for record deletion and apply them to our selection of pre-processed datasets (described in \autoref{sub:datasets}). Afterwards, we evaluate the impact of the simulated usage of the right to erasure on the performance of different classification methods as described in \autoref{sec:methodology}. As a result, we are able to compare implications of varying scenarios regarding the right to erasure.


\subsection{Experimental Setup}

Measuring performance differences in classification of an original dataset and a dataset with simulated use of the right to erasure requires selecting and pre-processing a dataset, designing and implementing specific simulated deletion scenarios and choosing and applying a machine learning algorithm. Consequently, our approach consists of four steps: pre-processing, baseline measurement, deletion and performance measurement.

\paragraph{Pre-processing} The dataset needs to be pre-processed before the machine learning algorithm uses it as training and test data to perform the classification task. During the pre-processing step, we remove unused columns, define generalisation hierarchies for every attribute of the dataset and perform one-hot encoding of categorical QIDs. 

\paragraph{Baseline calculation} After pre-processing, the baseline results on the data without further manipulation is calculated. This baseline (as well as a zero-rule baseline) are necessary to assess the impact of data record deletion on machine learning performance. The data is randomly split into training (70\%) and test data (30\%). The classifiers are trained using the training data and the resulting model is used to predict the target variable on the test data records.

\paragraph{Deletion}
In order to measure the machine learning performance difference, a specified number of records in the dataset are removed following different scenarios and assumptions. The percentage of records removed as well as the choice of scenario are configurable before the experiment begins.

\paragraph{Performance measurement}
The last step consists of retrieving the four performance measures for the classification tasks -- classification accuracy~($Acc$), precision~($Prec$), recall~($Rec$) and F\textsubscript{1}~score -- and comparing them to the results from the baseline calculation.
The measures are defined in terms of number of true positives~($TP$), true negatives~($TN$), false positives~($FP$) and false negatives~($FN$) as follows:
\begin{align*}
    Acc = \frac{TP+TN}{TP+TN+FP+FN}
\end{align*}
\begin{align*}
    Prec = \frac{TP}{TP+FP} \quad Rec = \frac{TP}{TP+FN} \quad F_1 &= 2 \frac{P \cdot R}{P+R}
\end{align*}

We conduct various different experiments via the parameters dataset, deletion percentage margin, deletion method, incremental mode and machine learning algorithm.

The original dataset is pre-processed and used in conjunction with the ML algorithm to establish the baselines. The experiment performs the chosen deletion method (either deleting records entirely at random or deleting records with certain attribute values with a higher probability) on the pre-processed dataset. The deletion is either done from scratch for each intended record deletion percentage, or alternatively (in ``incremental mode'') deletion is performed incrementally. Afterwards, the ML algorithm is used on the resulting smaller dataset and performance measures are determined and compared to the baseline numbers.

We implemented a command-line tool utilising the Python programming language to perform our experiments. The experiment parameters are specified via command line arguments, while the required options for pre-processing the dataset are configured inside the main experiment file. The datasets are stored in the \texttt{csv} format inside the dataset folder structure. Our implementation is open-source and available for further development and research.

\subsection{Deletion Methods}
\label{sub:deletion_methods}
To be able to examine different kinds of scenarios (e.\,g.\ those in which records with specific attributes are more likely to be removed due to different likelihoods of using the right to erasure), we developed different methods of applying a bias depending on specific value attributes. We ensure the comparability of our results despite nominally ``random'' deletion by utilising a fixed seed for the random number generator.

\paragraph{Random}
This mode randomly chooses as many records of the dataset as specified by the deletion percentage and removes them. No bias is applied; this mode assumes the likelihood of using of the right to erasure is uniform across all demographic and other cleavages. This is the standard mode used for all attributes not specifically mentioned in the rest of this section.

\paragraph{Selection}
One or several values of an attribute are assumed to increase the chances of the according entry being removed. The probability of deletion for records with these values is twice as high as for other records. This deletion mode is applied for the \texttt{salary-class} and \texttt{marital-status} attributes of \textsc{adult} as well as for the attribute \texttt{severity} of \textsc{mgm}  in \autoref{sec:evaluation}.

\paragraph{Thirds}
For numeric attributes with many different values, this mode can be applied to split the dataset into three probability groups. The values are split into thirds by using $\sfrac{1}{3}$ and $\sfrac{2}{3}$ percentiles and are weighted accordingly; the group with the smallest values has the standard chance of removal, while the probability for deletion is twice and thrice as high for the median and high values group, respectively. This deletion mode is used with the \texttt{median\_house\_value} and \texttt{latitude} attributes of \textsc{cahousing} in~\autoref{sec:evaluation}. Reversed possibilities are applied to the attribute \texttt{longitude}.

\paragraph{Age}
We apply this mode, which works similarly to the selection mode, to age attributes. Every record with an age value ${<}45$ has twice the probability of erasure as those with an age value ${\geq}45$. Although there is only limited research on age distributions of people actively protecting their privacy, a survey by Cisco~\cite{cisco2019cps} found that the majority of ``Privacy Actives'' (i.\,e.\ people who care about, are willing to act on or have acted to increase their own privacy) are aged between 18 and 44. Therefore, we chose the age 45 to be the cut-off between the two probability groups for this deletion scenario. Age mode is used in~\autoref{sec:evaluation} for the attributes \texttt{age} and (in combination with selection mode) for \texttt{age}+\texttt{marital-status} of\textsc{adult}.

\paragraph{Positive numeric}
This mode is intended for numeric attributes with only few different values or originally categorical attributes that were sorted and substituted by a corresponding numeric value beforehand. Deletion scenarios utilising this mode assume that records with a high value of the chosen attribute (in our use cases, this corresponds to higher likelihood of a privacy-critical attribute or higher socioeconomic status) are more likely to use the right to erasure. The mode scales the deletion probabilities linearly with increasing numeric values, i.\,e.\ for an attribute value of five the deletion probability is five times as high as for a value of one. It is also possible to reverse the deletion probability correlations with attribute values and to conversely assign higher possibilities to smaller values. This mode is used for the attributes \texttt{contraceptive\_method}, \texttt{wife\_edu} and  \texttt{num\_children} of \textsc{cmc} as well as the attributes \texttt{shape} and \texttt{bi\_rads\_assessment} of \textsc{mgm} in~\autoref{sec:evaluation}, whereas reversed possibilities are applied to the attribute \texttt{ocean\_proximity} of \textsc{cahousing}.

\subsection{Incremental Mode}
\label{sub:incremental}
Although deleted records are held constant for the same deletion percentage value in different runs of our experiment tool (by using fixed seeds for the random number generator), the deleted records of different percentage values in the same run are independent. The optional incremental mode can be used in conjunction with all other modes and changes this behaviour: Records deleted for a specific deletion percentage are also removed for the subsequent deletion percentage, with as many ``new'' records additionally deleted as is required to match the target deletion percentage.



\subsection{Datasets}
\label{sub:datasets}
In the experiments, four real-world datasets are used: \textsc{adult},\footnote{\url{https://www.kaggle.com/uciml/adult-census-income}} \textsc{cahousing},\footnote{\url{https://www.kaggle.com/camnugent/california-housing-prices}} \textsc{cmc}\footnote{\url{https://archive.ics.uci.edu/ml/datasets/Contraceptive+Method+Choice}} and \textsc{mgm}.\footnote{\url{https://archive.ics.uci.edu/ml/datasets/Mammographic+Mass}} 
We summarise the main characteristics in \autoref{table:datasets}.\footnote{Note that compared to the original \textsc{cahousing} dataset, we merged the very sparsely populated classes \texttt{NEAR BAY} and \texttt{ISLAND} into \texttt{NEAR OCEAN} to achieve a sensible value split.}

Since we conduct our experiments on the same datasets used in our last paper, we use the same generalisation hierarchies for all datasets as in our previous study~\cite{slijepcevic2021kanonymity}.
We repeat the details for two datasets here: For \textsc{adult}, the hierarchies are the same as in~\cite{Prasser2014}. For \textsc{cahousing}, the generalisation hierarchies for longitude and latitude were created through the usual step-wise reduction in precision by removing trailing digits, whereas the remaining generalisation hierarchies were chosen through division by equidistant boundaries (except for median income, where we chose a larger interval for the largest values).
\section{Evaluation}
\label{sec:evaluation}

\begin{figure*}[!htbp]
    \centering
    \includegraphics[width=\linewidth]{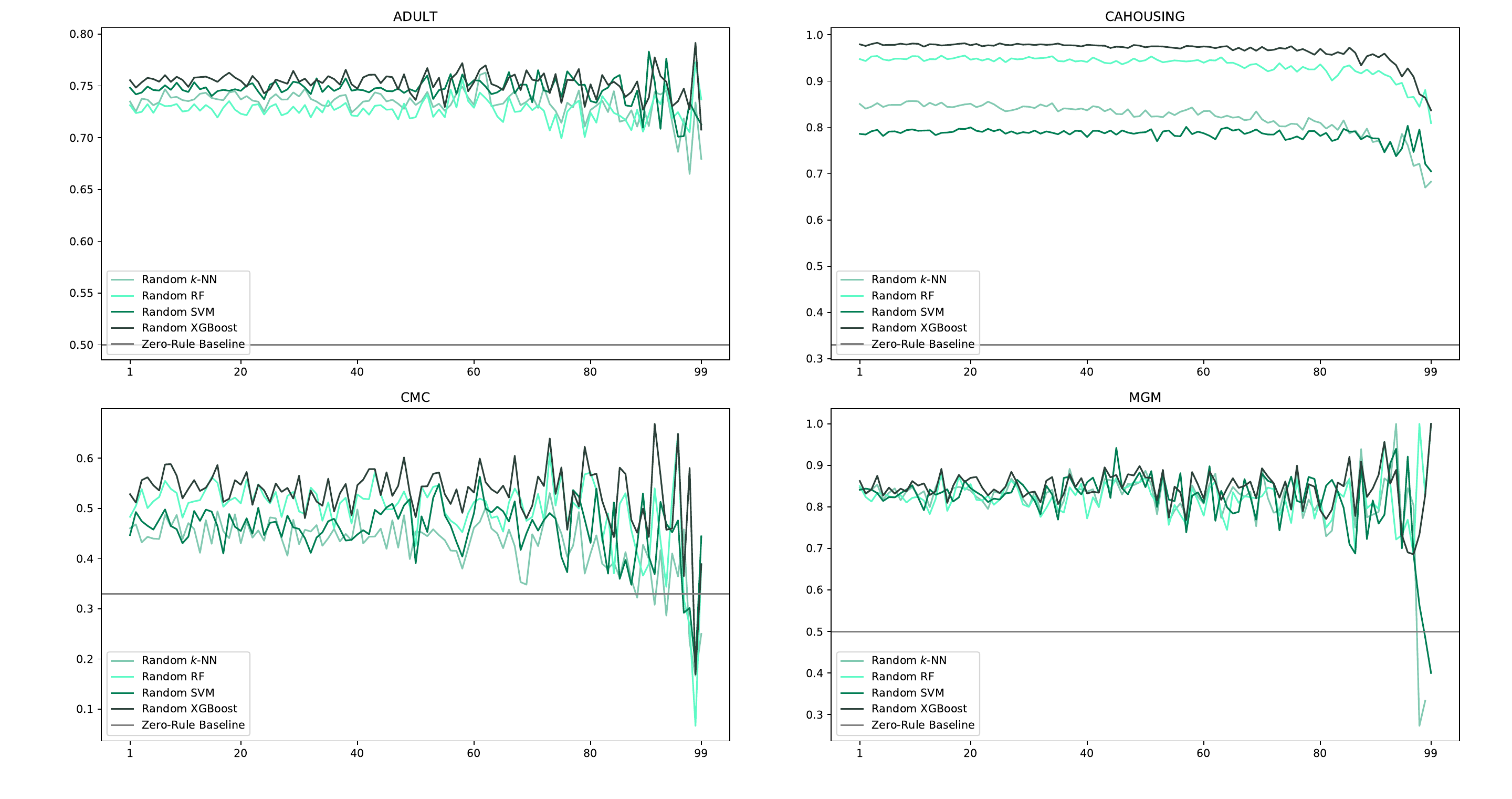}
    \caption{Overview of F\textsubscript{1}~scores for random deletion for all datasets and classifiers.}
    \label{fig:classifier_overview}
\end{figure*}

As described in~\autoref{sec:design}, we conducted experiments consisting of different datasets, machine learning algorithms and deletion modes. \autoref{fig:classifier_overview} shows the performance of the classification task in terms of F\textsubscript{1}~score for the random deletion mode and the classifiers $k$-NN, RF, SVM and XGBoost.

The random deletion mode chooses the configured amount of records at random and the records selected can vary for subsequent runs with higher deletion percentages, as explained in \autoref{sub:deletion_methods}. As a consequence, the classification performance features high degrees of fluctuation and consecutive runs with increasing deletion percentages can result in rather different F\textsubscript{1}~scores.

While the random deletion mode shows strong fluctuations across all datasets, the impact on ML performance appears to be less noticeable on average for larger datasets such as \textsc{adult} and \textsc{cahousing} when compared to smaller datasets, i.\,e.\ \textsc{cmc} and \textsc{mgm}.

In our previous study~\cite{slijepcevic2021kanonymity},
we found that XGBoost performed best on average across the different datasets in terms of F\textsubscript{1}~score. As shown in \autoref{fig:classifier_overview}, this is also the case for the deletion experiments in this study. The F\textsubscript{1}~score results of XGBoost show the same strong fluctuations as the other classifiers.

Although the results for higher deletion percentages are not of practical relevance, we include them nonetheless to give a comprehensive account of our results.

\begin{figure*}[!htbp]
    \centering
    \includegraphics[width=\linewidth]{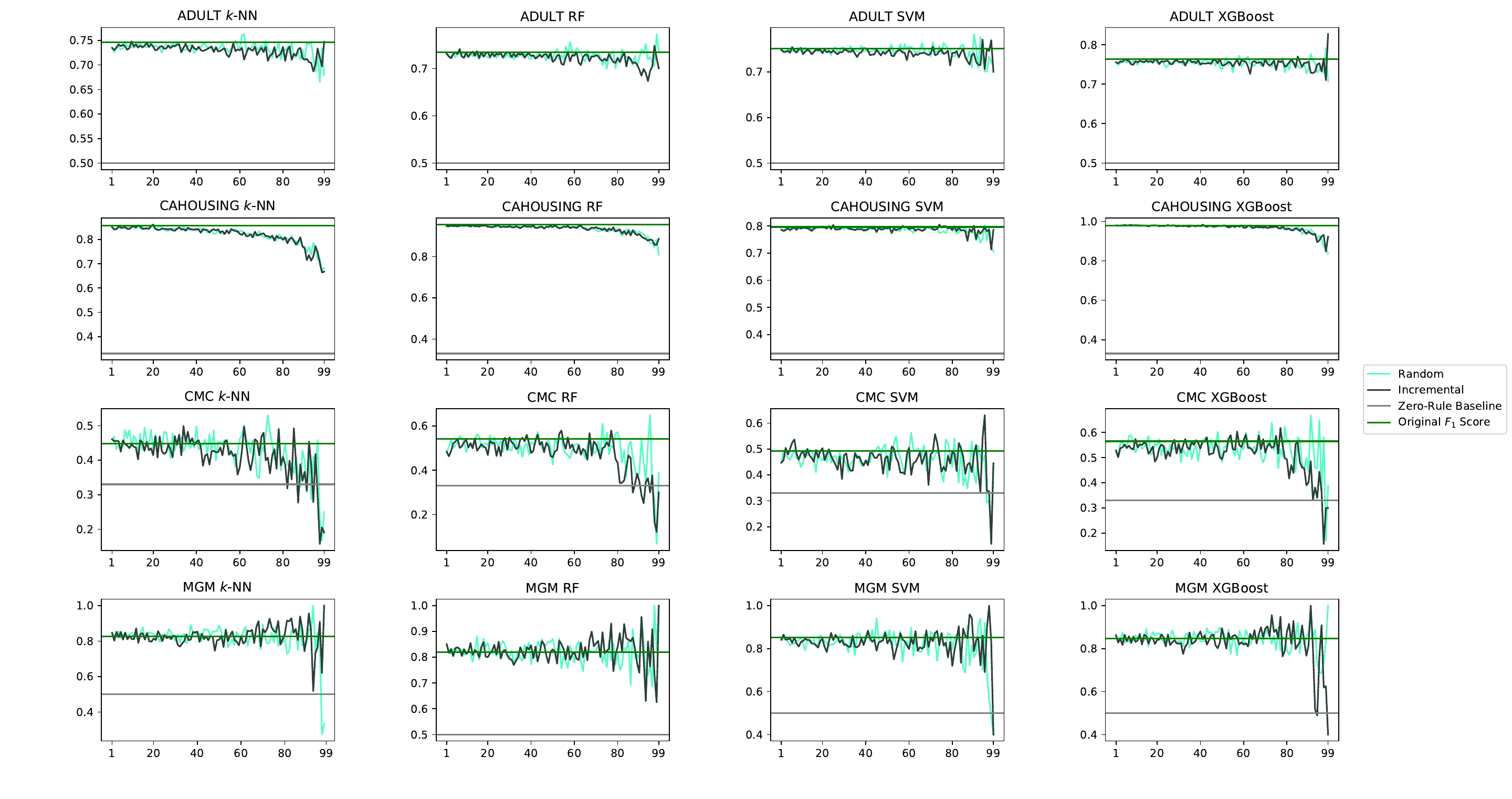}
    \caption{Overview of F\textsubscript{1}~score difference between random and incremental deletion for all datasets and classifiers.}
    \label{fig:random_incremental}
\end{figure*}

In addition to the standard mode of selecting all records at random for every deletion percentage, we also implemented the so-called ``incremental'' deletion mode, which deletes the records already selected for the preceding (lower) deletion percentage plus the required amount of additional records to achieve the desired new deletion percentage. We compared the results of this deletion mode with those of random mode in \autoref{fig:random_incremental}. While there are some cases where the incremental mode performs increasingly worse over consecutive deletion percentages, as shown for \textsc{cmc} with both RF and XGBoost, the opposite occurs as well, e.\,g.\ for \textsc{adult} with SVM between deletion percentages $70$\,\% and $80$\,\%. Contrary to our expectations (of a more noticeable downward trend in incremental deletion mode results), \autoref{fig:random_incremental} indicates that increasing the amount of deleted records (while respecting previously deleted records) does not necessarily have a significant impact on ML performance. 

Another experiment concerns the different impacts of the purely random deletion mode and the deletion modes prioritising records based on specific attribute values (as described in \autoref{sub:deletion_methods}). We applied the ``selection'' deletion method for the \texttt{salary-class} attribute of the \textsc{adult} dataset and the \texttt{severity} attribute of the \textsc{mgm} dataset. In this scenario, we assumed that wealthier people were more likely to make use of the right to erasure and assigned each record with the \texttt{salary-class} value \texttt{>50K} a higher deletion probability. Similarly, records with the \texttt{severity} value $1$ (malign) of the \textsc{mgm} dataset were assigned a higher probability, based on the assumption that people with more severe diagnoses were more likely to want their information deleted. The target variables \texttt{ocean\_proximity} and \texttt{contraceptive\_method} of their respective datasets \textsc{cahousing} and \textsc{cmc} apply the ``positive numeric'' deletion method, meaning that people living closer to the ocean and people using short-term contraceptive methods, respectively, were assigned the highest probabilities of using the right to erasure.

\begin{figure*}[!htbp]
    \centering
    \includegraphics[width=\linewidth]{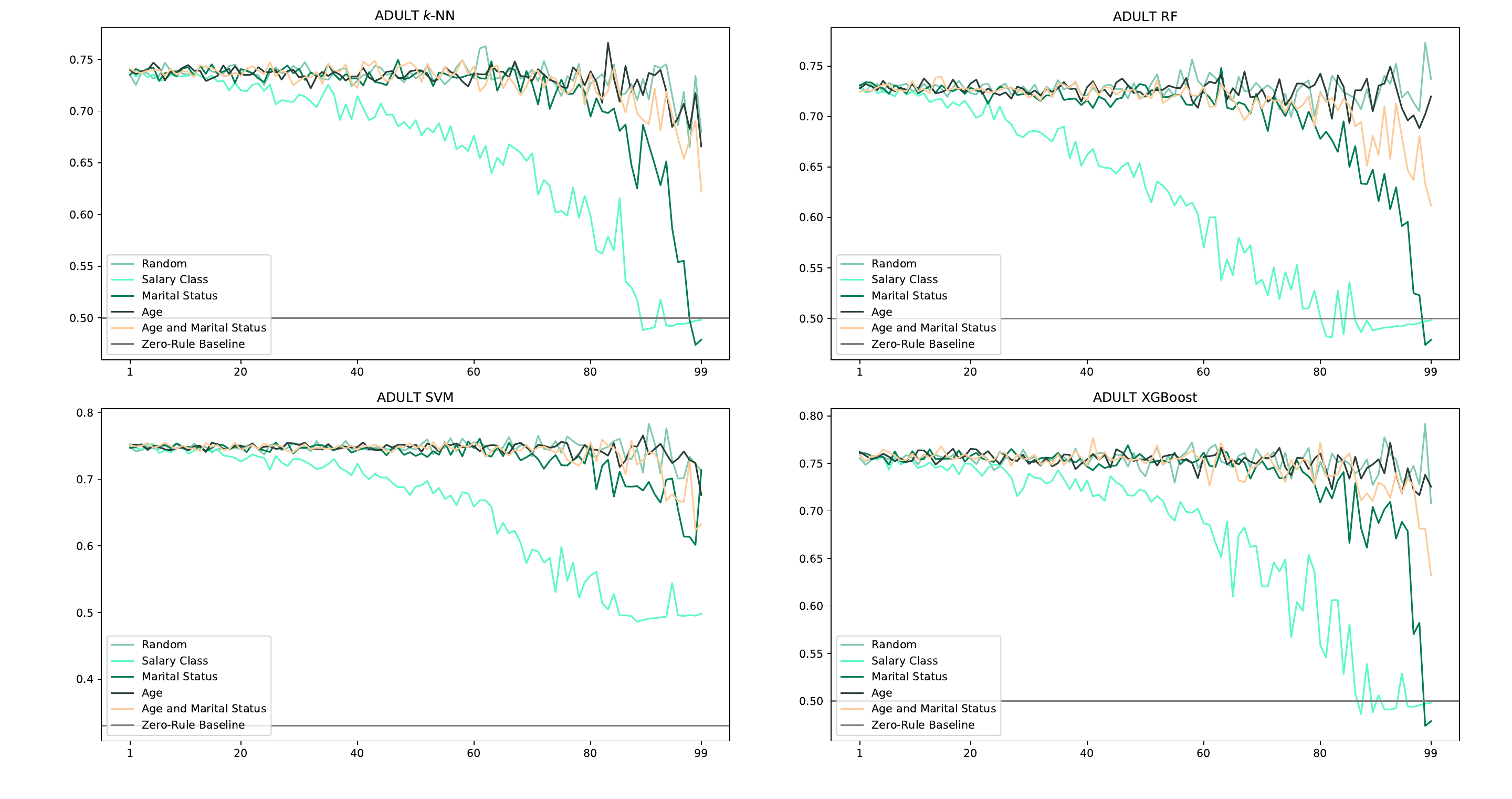}
    \caption{Comparison of F\textsubscript{1}~scores for differently biased deletion for all classifiers on the \textsc{adult} dataset.}
    \label{fig:bias_adult}
\end{figure*}

\begin{figure*}[!htbp]
    \centering
    \includegraphics[width=\linewidth]{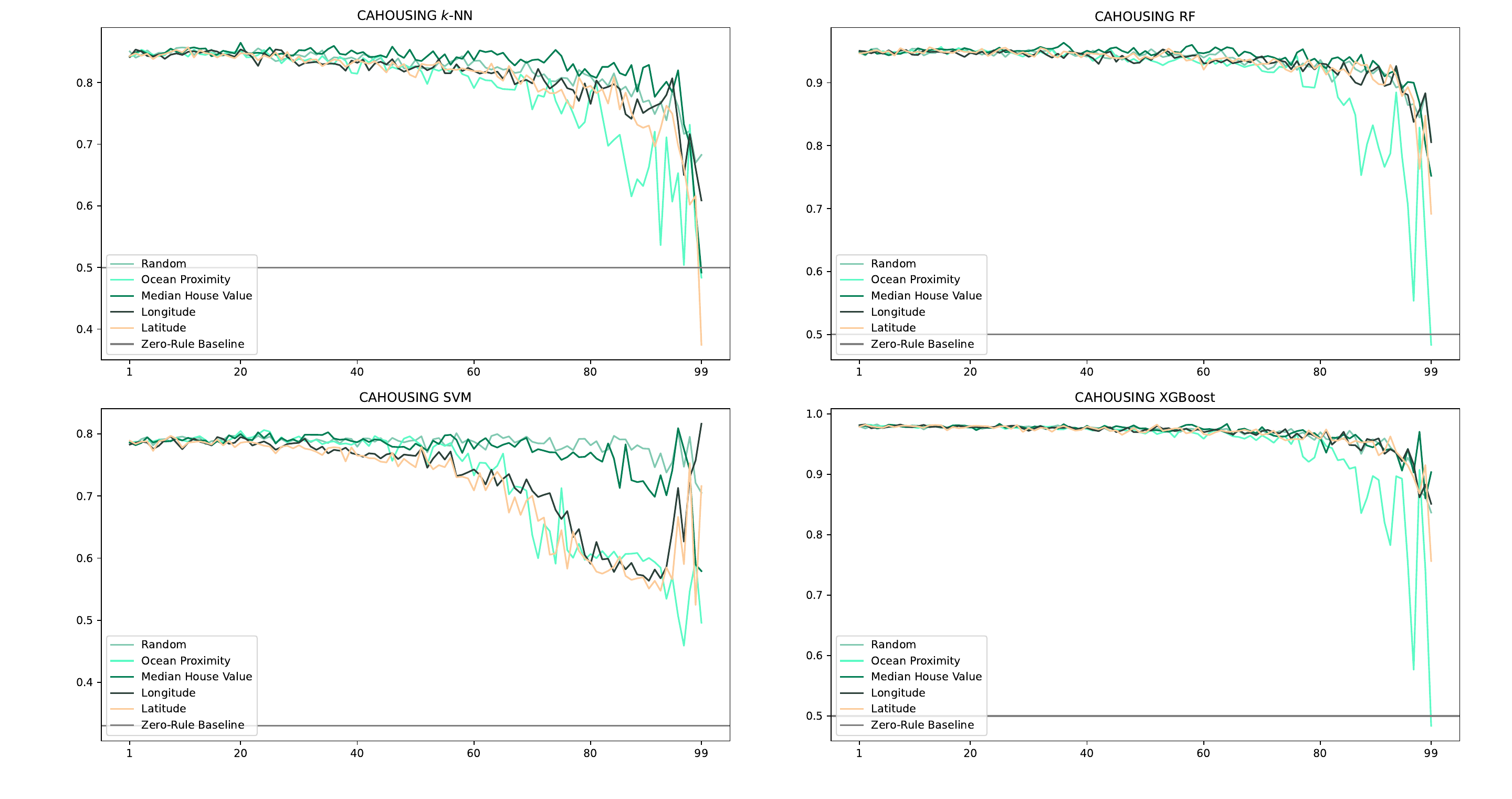}
    \caption{Comparison of F\textsubscript{1}~scores for differently biased deletion for all classifiers on the \textsc{cahousing} dataset.}
    \label{fig:bias_cahousing}
\end{figure*}

\begin{figure*}[!htbp]
    \centering
    \includegraphics[width=0.94\linewidth]{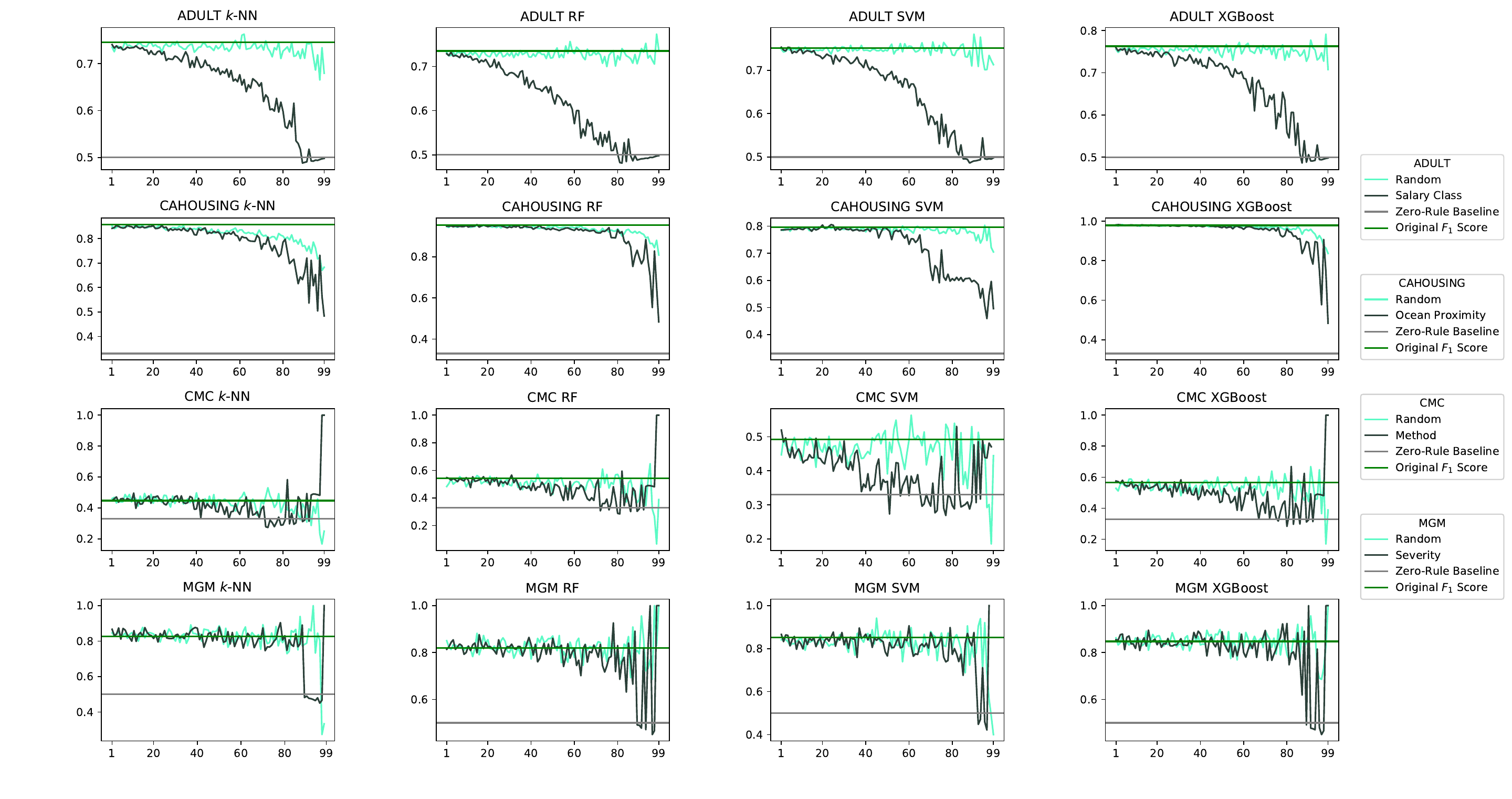}
    \caption{Overview of F\textsubscript{1}~score difference between random and biased deletion for all datasets and classifiers. The bias was set on the target variable of the classification.}
    \label{fig:random_target}
\end{figure*}

\begin{figure*}[!htbp]
    \centering
    \includegraphics[width=\linewidth]{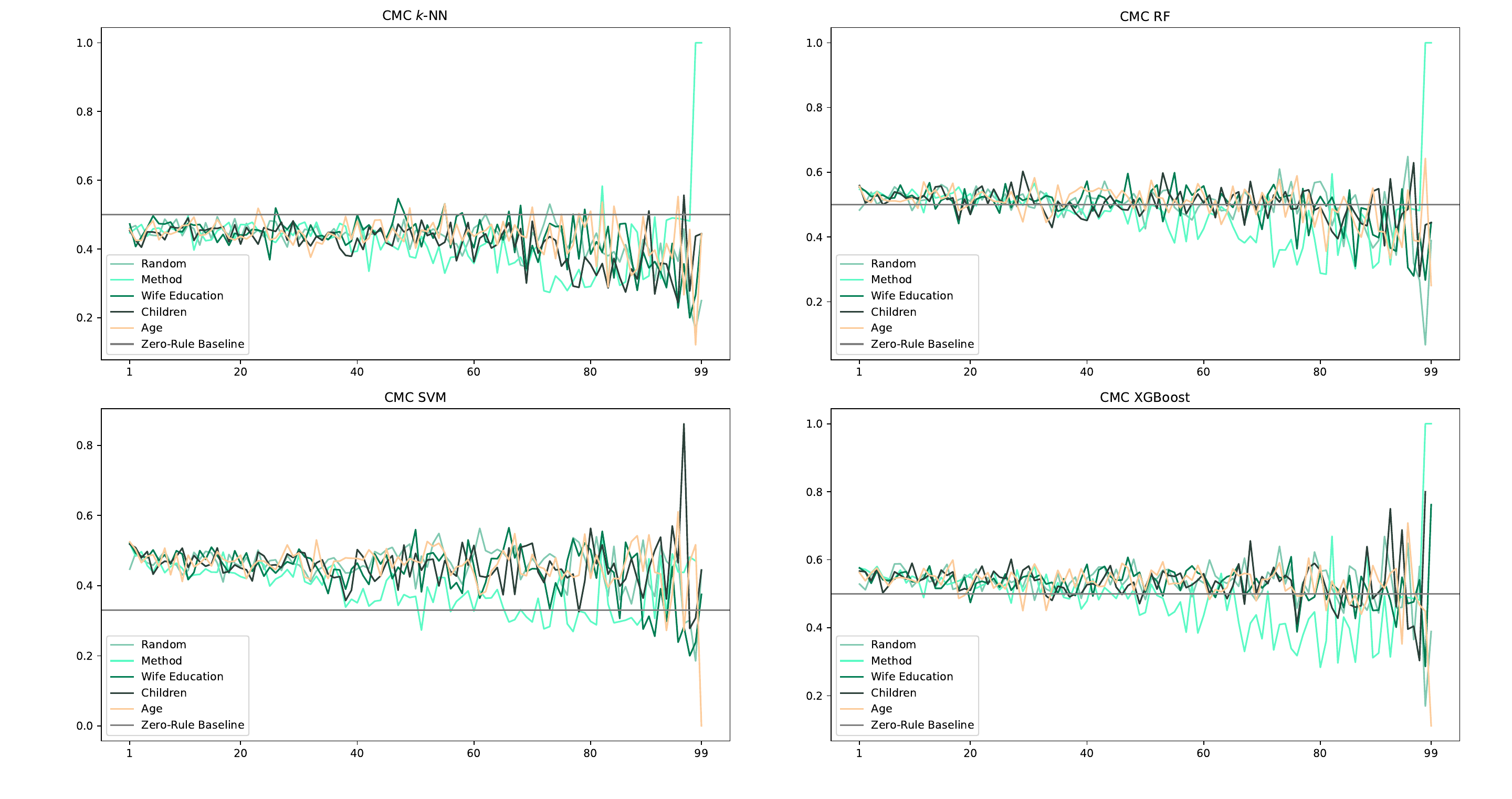}
    \caption{Comparison of F\textsubscript{1}~scores for differently biased deletion for all classifiers on the \textsc{cmc} dataset.}
    \label{fig:bias_cmc}
\end{figure*}

\begin{figure*}[!htbp]
    \centering
    \includegraphics[width=\linewidth]{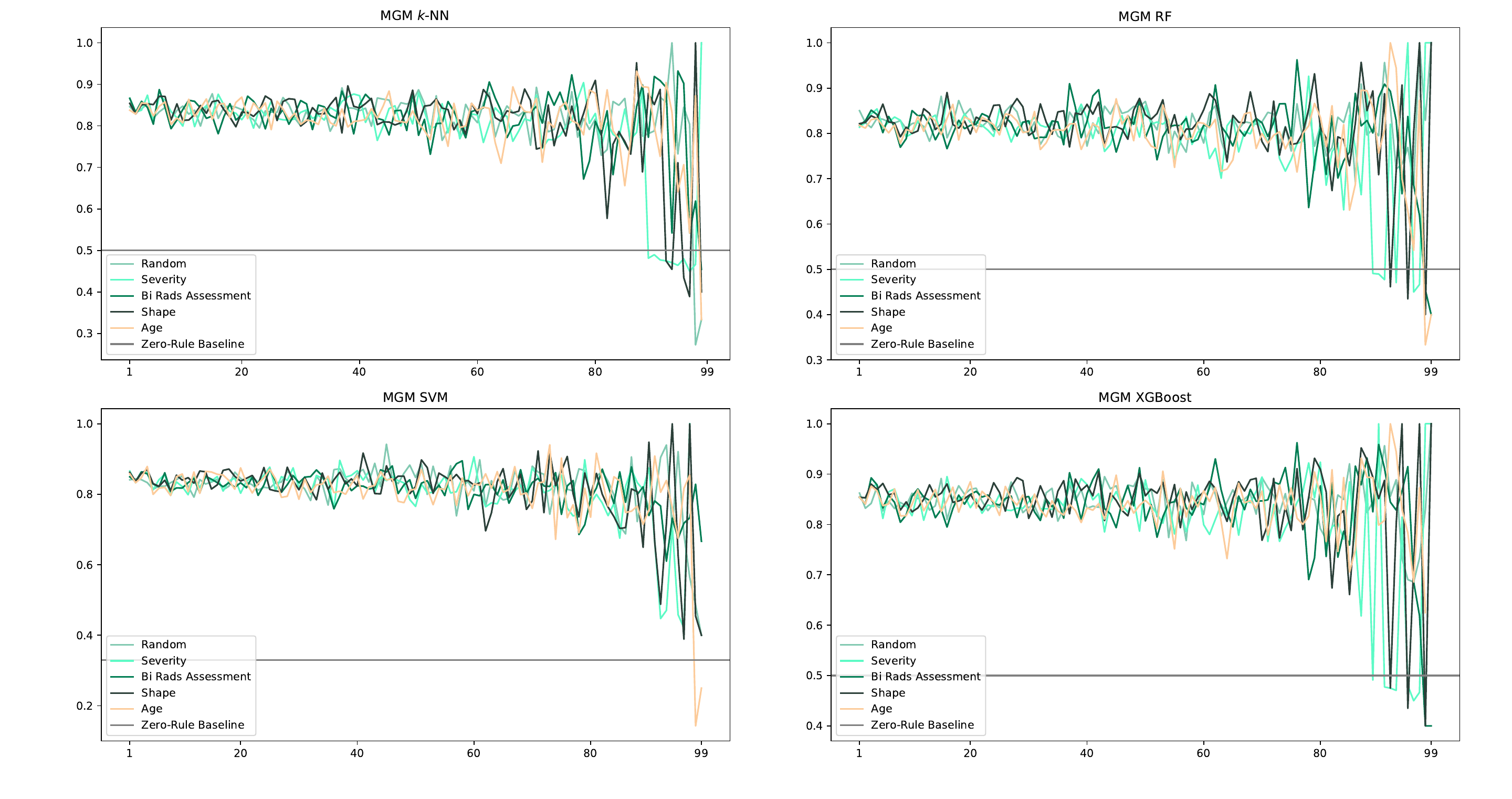}
    \caption{Comparison of F\textsubscript{1}~scores for differently biased deletion for all classifiers on the \textsc{mgm} dataset.}
    \label{fig:bias_mgm}
\end{figure*}

The results of this experiment show that biased deletion methods applied to the target variable of the corresponding dataset have a more noticeable impact on ML classification performance in contrast to the random deletion method. The F\textsubscript{1}~score difference between the two deletion modes grows with increasing deletion percentages; moreover, this effect seems to be even stronger for larger datasets (such as \textsc{adult} and \textsc{cahousing}). Hence our findings imply that in case a relevant fraction of people with the same or similar target variable attribute values (or similar values for attributes which strongly influence the target variable in terms of ML classification) use their right to erasure, there will be a strong negative impact on ML classification performance. However, this effect seems to remain mostly negligible for deletion percentages up to approximately 20\,\% (on average) in our experiments, as is evident from the bias comparisons in~\cref{fig:bias_adult,fig:bias_cahousing,fig:bias_cmc,fig:bias_mgm} or the overview in~\autoref{fig:random_target}.

To further determine the impact of different attributes and according attribute values of datasets on the ML classification performance, we performed several biased deletion mode experiments with different biases. We included the results for random deletion mode as well as for biased deletion mode for the target variable for each dataset in the figures for reasons of clarity and comprehensibility, and added three other kinds of biases. \autoref{fig:bias_adult} compares the different biases applied to the \textsc{adult} dataset. The bias ``Age'' applies the age deletion method as described in~\autoref{sub:deletion_methods}. ``Marital Status'' applies the selection deletion method focusing on the attribute \texttt{married-civ-spouse}. ``Age and Marital Status'' combines both methods resulting in four possible combinations (from high to low likelihood): young and married, young and unmarried, old and married as well as old and unmarried. Our results show that across all classifiers, the ``Marital Status'' bias has a strong influence on ML performance, while the ``Age'' bias only has minor impact. The F\textsubscript{1}~scores for ``Age and Marital Status'' confirm these findings, as the results lie between the ``Age'' and ``Marital Status'' biases.

In contrast to the results for \textsc{adult}, our \textsc{cahousing} experiments feature more varying F\textsubscript{1}~scores for the same bias utilising different classifiers, as depicted in \autoref{fig:bias_cahousing}. While for $k$-NN, RF and XGBoost the ``Ocean Proximity'' bias performs worst, especially for higher deletion percentages, the biases ``Longitude'' and ``Latitude'' have the highest negative impact for most deletion percentages when used in conjunction with the SVM classifier. Another interesting difference can be seen in the increased F\textsubscript{1}~scores of the ``Median House Value'' bias compared to the random deletion mode, which are evident for $k$-NN and RF as well as (although not as clearly) for XGBoost. The additional biases ``Longitude'', ``Latitude'' and ``Median House Value'' use the thirds deletion method as explained in \autoref{sub:deletion_methods}.``Latitude'' and ``Median House Value'' assign higher probabilities to higher values, hence people living further to the north or living in more expensive houses are more likely to use their right to erasure in our scenario. ``Longitude'' assigns higher probabilities to lower values, thus prioritising people living further to the west.

\begin{figure*}[!htbp]
    \centering
    \includegraphics[width=\linewidth]{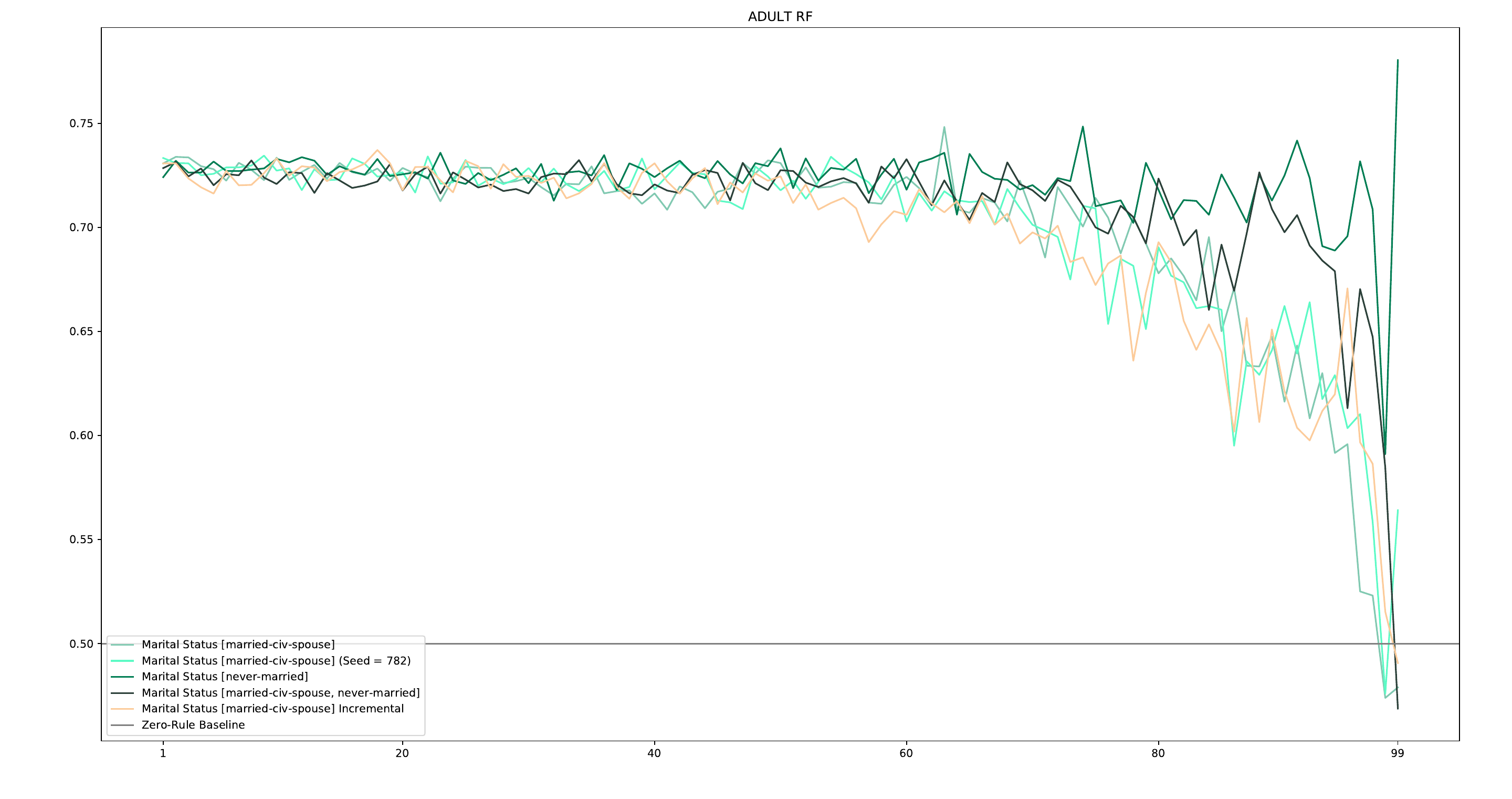}
    \caption{Comparison of F\textsubscript{1}~scores for deletion method variations of the same attribute for all classifiers on the \textsc{adult} dataset.}
    \label{fig:adult_married}
\end{figure*}
We performed another bias comparison on the \textsc{cmc} dataset utilising the ``Wife Education'', ``Children'' and ``Age'' biases. ``Age'' uses the age deletion method, while ``Wife Education'' and ``Children'' use the positive numeric deletion method. As a consequence, in these scenarios, highly educated women and families with more children are assigned higher deletion probabilities. None of the aforementioned biases appear to have a significant or noticeably increasing impact on ML classification performance (as shown in \autoref{fig:bias_cmc}). Since focusing the deletion on the target variable (as is the case for the ``Method'' bias) does not heavily degrade the performance either, the weaker impact when compared to the results for the \textsc{adult} and the \textsc{cahousing} datasets might be caused by the far smaller number of data records.


We conducted a bias experiment on the \textsc{mgm} dataset as well, using the biases ``Bi Rads Assessment'', ``Shape'' and ``Age''. The ``Age'' bias uses the age deletion method, whereas the other biases apply the positive numeric deletion method: The more malign the BI-RADS assessment or the more irregular the shape, the higher the possibility of the use of the right to erasure. Similar to the results for \textsc{cmc}, no bias has a significant effect on the highly fluctuating ML classification performance (as apparent in \autoref{fig:bias_mgm}). The impact is even less noticeable when compared to the \textsc{cmc} experiment, except at very high deletion percentages (approximately above 90\,\%). At higher percentages, deletion of the target variable has the strongest effect. These results support the conclusion that the weaker impact is caused by the smaller number of records, since the \textsc{mgm} dataset contains even fewer records than the \textsc{cmc} dataset.

\begin{figure*}[!htbp]
    \centering
    \includegraphics[width=0.94\linewidth]{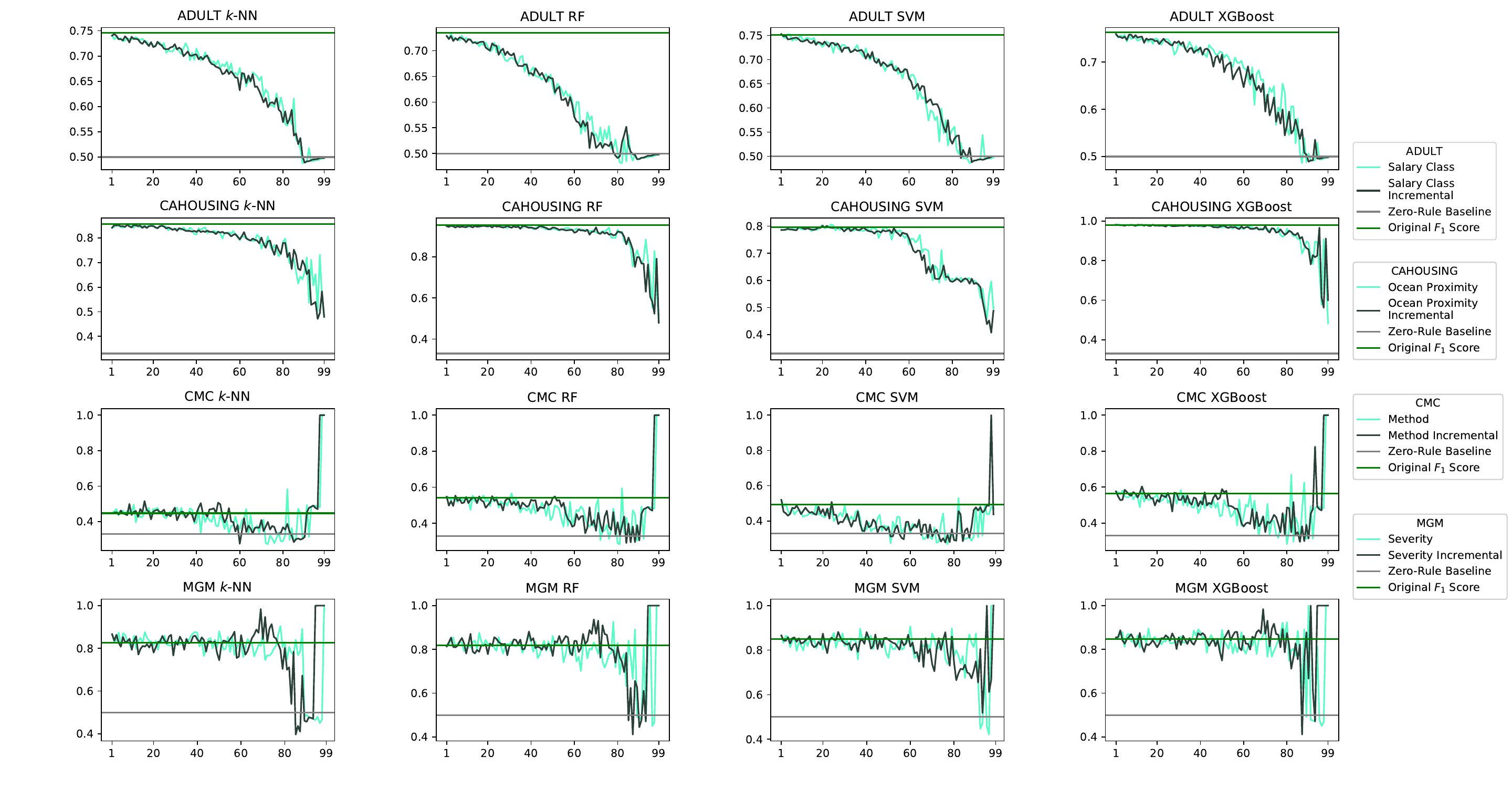}
    \caption{Overview of F\textsubscript{1}~score difference between biased and incremental deletion on the target variable for all datasets and classifiers.}
    \label{fig:inc_target_var}
\end{figure*}

Before conducting our experiments, we defined specific characteristics for the various deletion methods, such as the focus on specific attribute values or the use of a fixed seed for the random number generator. To determine the impact of these choices on our results, we performed another experiment utilising the ``Marital Status'' bias for the \textsc{adult} dataset with modified parameters. \autoref{fig:adult_married} shows our results for different deletion methods. ``Marital Status [married-civ-spouse]'' focuses on the same attribute value and uses the same fixed seed as in \autoref{fig:bias_adult}. ``Marital Status [married-civ-spouse] (Seed = 782)'' and ``Marital Status [married-civ-spouse] Incremental'' also target the attribute value \texttt{married-civ-spouse}, with the former applying a different random seed and the latter using the incremental method. ``Marital Status [never-married]'' assigns higher percentages to adults who never married. Finally, ``Marital Status [married-civ-spouse, never-married]'' combines both biases by assigning the same increased probability to records with either of the attribute values \texttt{married-civ-spouse} or \texttt{never-married}. The results of ``Marital Status [never-married]'' and `Marital Status [married-civ-spouse, never-married]'' show that preferring only (or additionally) attribute values that are less important for the ML classification performance has a noticeably reduced impact. Although the bias utilising a different random seed seems to perform similar on average, the fluctuations are quite different. As already apparent in \autoref{fig:random_incremental} as well as \autoref{fig:inc_target_var}, the bias ``Marital Status [married-civ-spouse] Incremental'' features only a slight negative influence on the ML classification performance for higher deletion percentages.

\section{Future Work}
\label{future work}

Our work could be expanded upon in several directions. For one, repeating our experiments with various other datasets, especially real-world data, could drastically increase the insight into what impact different types of deletion behaviour might have. Furthermore, including more datasets of different sizes could help prove or disprove our hypothesis that larger datasets suffer a more significant impact at higher deletion percentages than smaller ones.

Creating and utilising more deletion methods and biases could create a more holistic general view and provide additional information on the impact of different characteristics.

Finally, another possibility to vastly increase the quality of our results (although somewhat outside our scope) would be to conduct studies and surveys regarding the use of the right to erasure and investigate correlations with demographic, socioeconomic and other characteristics. Due to the lack of such studies in existing research literature, as mentioned in \autoref{sub:Right to Erasure and Privacy Awareness}, our experiments are based on what we consider plausible assumptions and somewhat arbitrary decisions. Retrieving data on which kind of people are more and less likely to make use of their right to erasure (and for what reason) would be a significant contribution to research conducted in this field.
\section{Conclusion}
\label{sec:conclusion}

Announcements of various security and privacy incidents, public awareness campaigns as well as other events have led to more and more users caring about and actively trying to protect their privacy. Additionally, since the GDPR has been in force since 25 May 2018 in the EU, more users have become aware of the right to erasure of the data stored about them by companies and organisations. As the use of this right might increase even more in the future, data deletion could impact the data quality for ML classification tasks.

We performed several experiments and applied various deletion patterns on different datasets to gain insight into the impact on data quality caused by the application of the right to erasure. Due to the lack of empirical research literature on the likelihood of different groups of people making use of their right, we defined different scenarios based on plausible assumptions. This enabled us to analyse the impact of deletion patterns focused on specific characteristics and then assess the negative effect of such data erasure on data quality.

Our results showed that the impact on ML classification performance highly depends on the number of deleted records, the specific characteristics of the dataset and which attribute values are most important for the classification. The F\textsubscript{1}~score heavily fluctuates across the different datasets and deletion methods, but the impact is only moderate on average in most cases (except for very high deletion percentages).

The ML performance for increasing deletion percentages shows a more noticeable downward trend for experiments on larger datasets (\textsc{adult} and \textsc{cahousing}) when compared to the much smaller datasets \textsc{cmc} and \textsc{mgm}. We suspect that this effect most likely depends on the size of datasets and suggest additional research in this direction.

\section*{Acknowledgements}
This research was funded by the Austrian Research Promotion Agency (FFG) COIN project 866880 ``Big Data Analytics''. The financial support by the Austrian Research Promotion Agency and the Federal Ministry for Digital and Economic Affairs is gratefully acknowledged.

\bibliographystyle{IEEEtran}
\bibliography{Bibliography}
\appendices
\onecolumn
\section{Complementary Plots}

\section{Smoothed Plots}

\begin{figure*}[!htbp]
    \centering
    \includegraphics[width=0.94\linewidth]{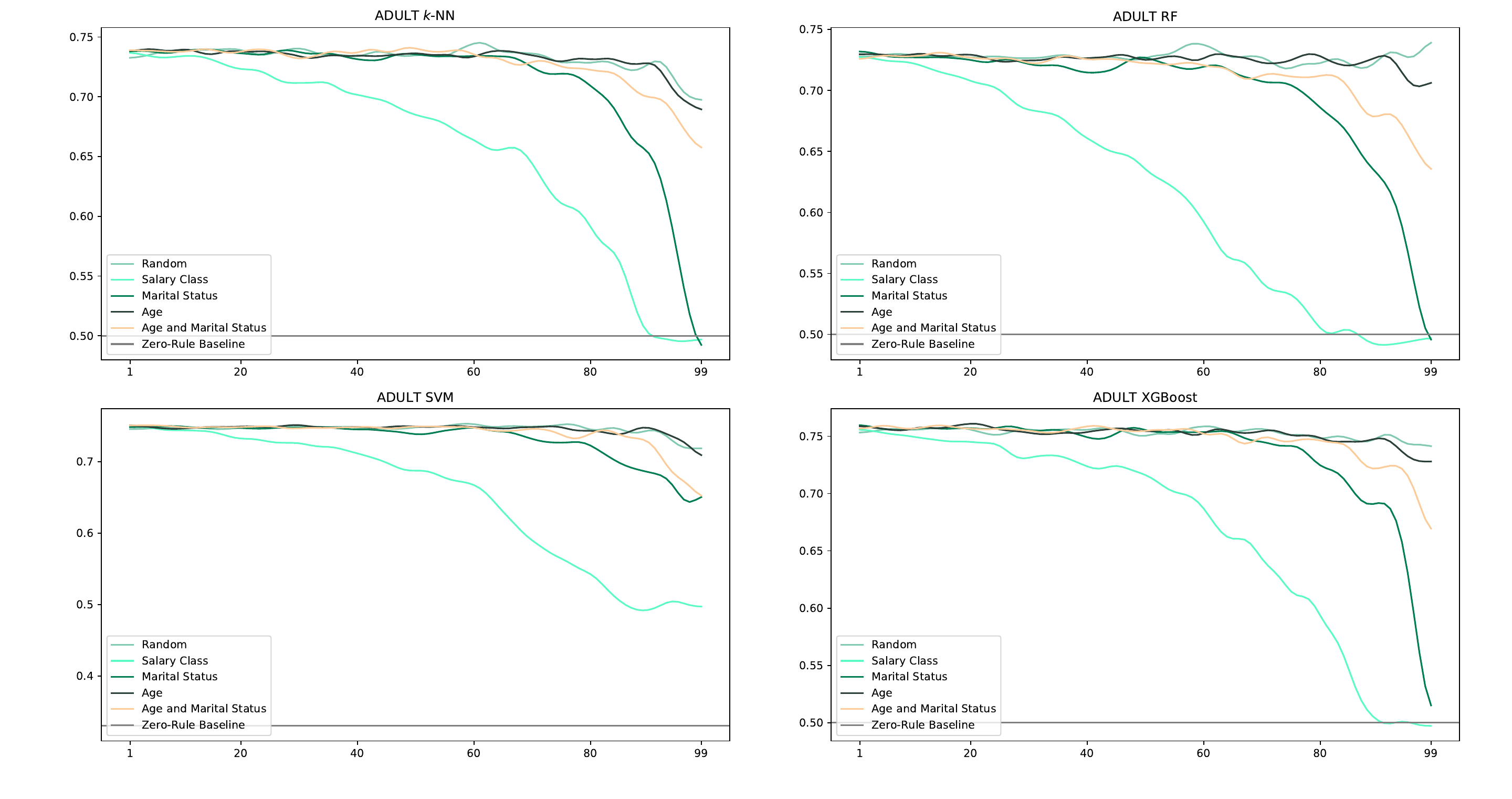}
    \caption{Comparison of F\textsubscript{1}~scores for differently biased deletion for all classifiers on the \textsc{adult} dataset. Scores are smoothed using a Gaussian filter with a $\sigma$ of 2.}
    \label{fig:bias_adult_smoothed_2}
\end{figure*}

\begin{figure*}[!htbp]
    \centering
    \includegraphics[width=0.94\linewidth]{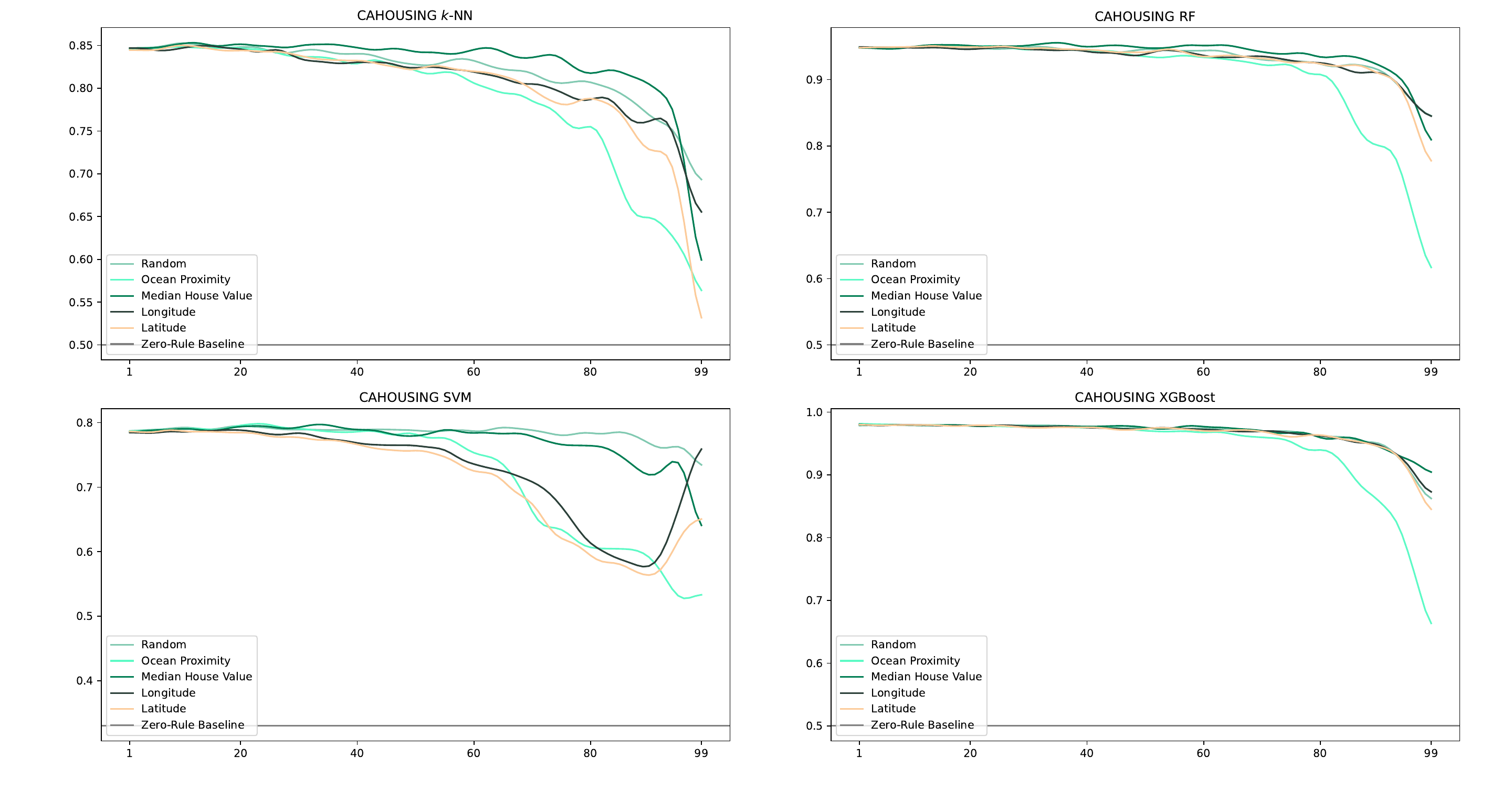}
    \caption{Comparison of F\textsubscript{1}~scores for differently biased deletion for all classifiers on the \textsc{cahousing} dataset. Scores are smoothed using a Gaussian filter with a $\sigma$ of 2.}
    \label{fig:bias_cahousing_smoothed_2}
\end{figure*}

\begin{figure*}[!htbp]
    \centering
    \includegraphics[width=\linewidth]{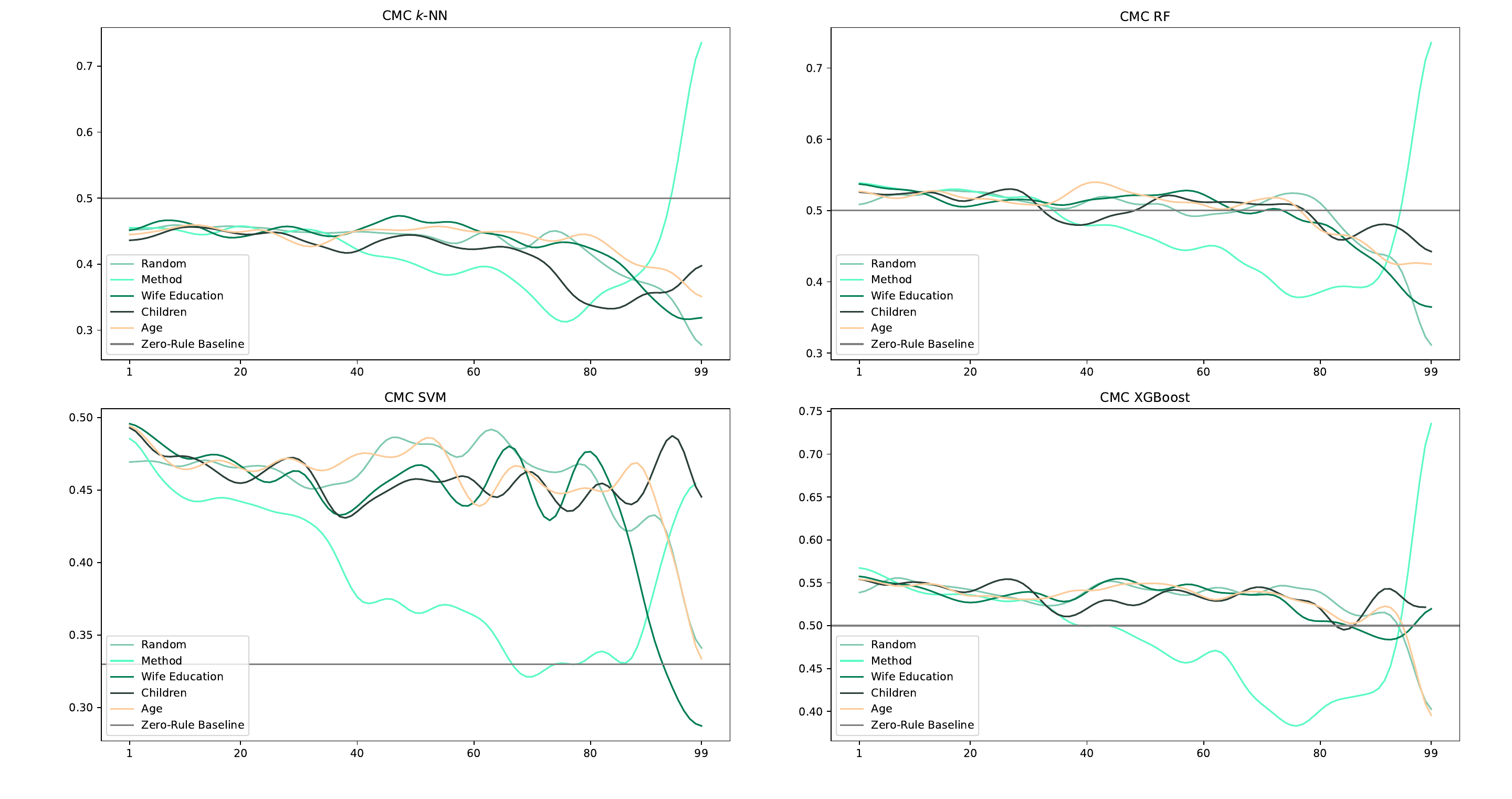}
    \caption{Comparison of F\textsubscript{1}~scores for differently biased deletion for all classifiers on the \textsc{cmc} dataset. Scores are smoothed using a Gaussian filter with a $\sigma$ of 3.}
    \label{fig:bias_cmc_smoothed_3}
\end{figure*}

\begin{figure*}[!htbp]
    \centering
    \includegraphics[width=\linewidth]{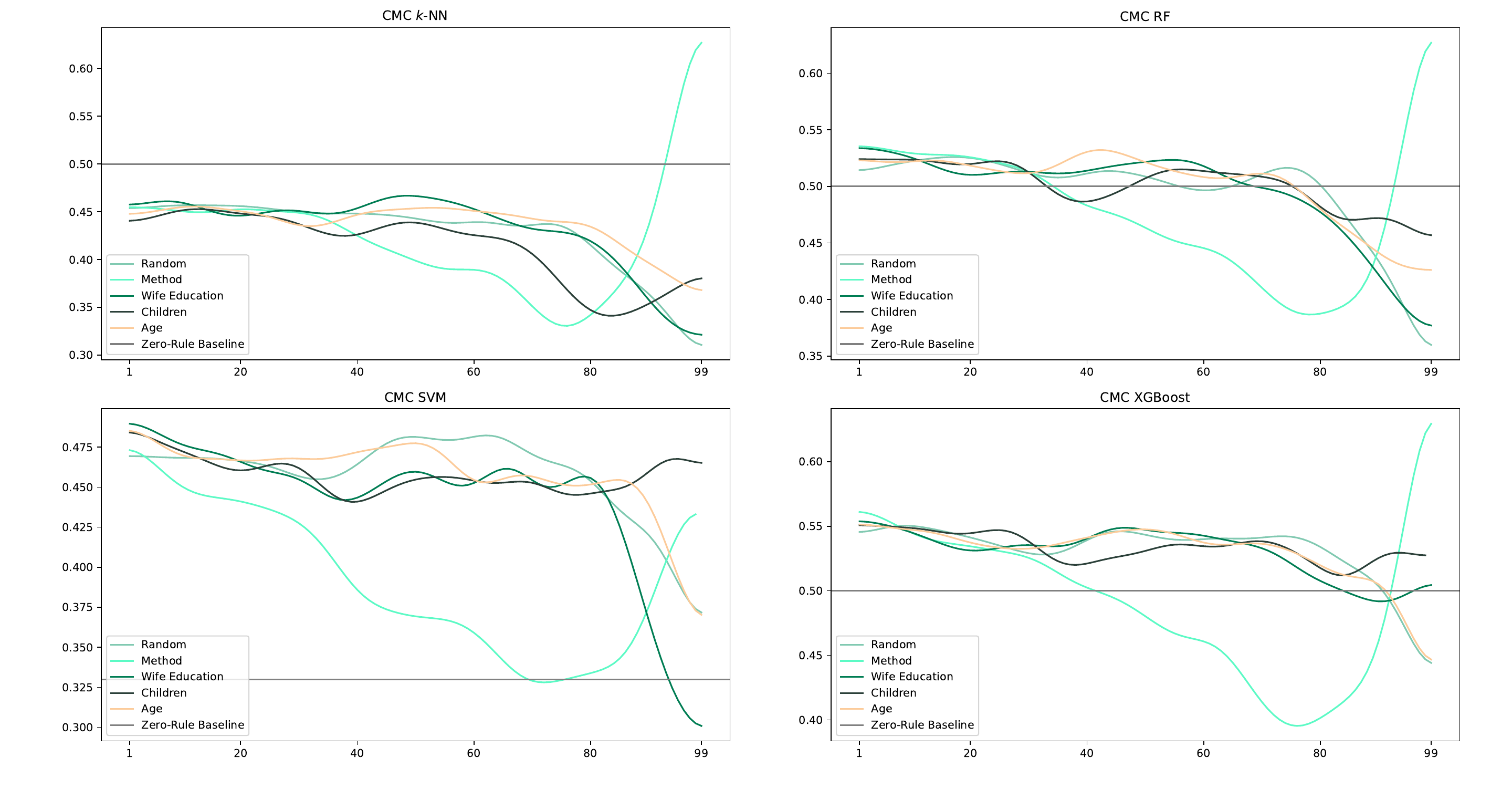}
    \caption{Comparison of F\textsubscript{1}~scores for differently biased deletion for all classifiers on the \textsc{cmc} dataset. Scores are smoothed using a Gaussian filter with a $\sigma$ of 5.}
    \label{fig:bias_cmc_smoothed_5}
\end{figure*}

\begin{figure*}[!htbp]
    \centering
    \includegraphics[width=\linewidth]{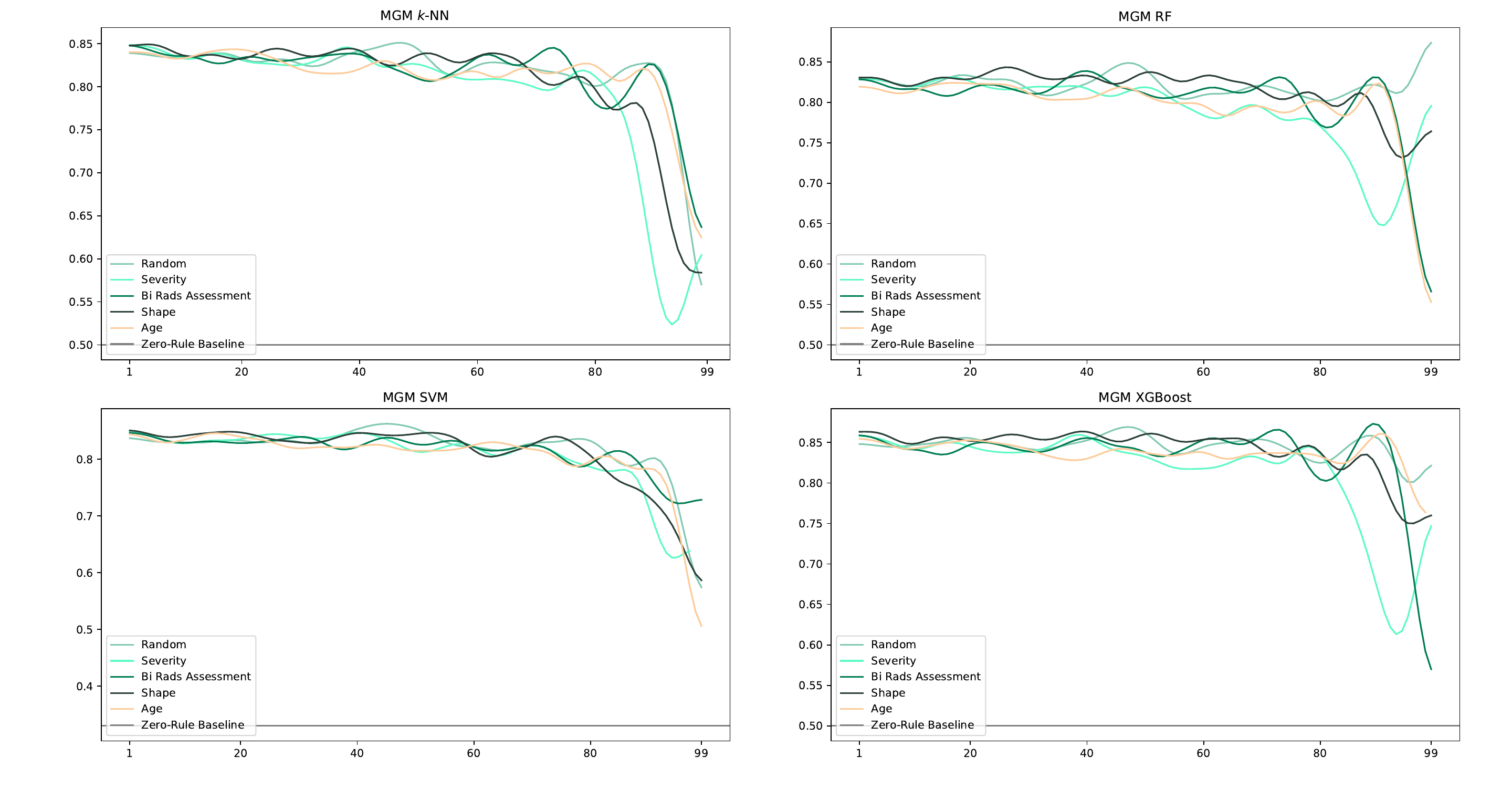}
    \caption{Comparison of F\textsubscript{1}~scores for differently biased deletion for all classifiers on the \textsc{mgm} dataset. Scores are smoothed using a Gaussian filter with a $\sigma$ of 3.}
    \label{fig:bias_mgm_smoothed_3}
\end{figure*}

\end{document}